# A New Adaptive Noise Covariance Matrices Estimation and Filtering Method: Application to Multi-Object Tracking

Chao Jiang, Zhiling Wang, Shuhang Tan, and Huawei Liang*

*Abstract-* **Kalman filters are widely used for object tracking, where process and measurement noise are usually considered accurately known and constant. However, the exact known and constant assumptions do not always hold in practice. For example, when lidar is used to track noncooperative targets, the measurement noise is different under different distances and weather conditions. In addition, the process noise changes with the object's motion state, especially when the tracking object is a pedestrian, and the process noise changes more frequently. This paper proposes a new estimation-calibration-correction closed-loop estimation method to estimate the Kalman filter process and measurement noise covariance matrices online. First, we decompose the noise covariance matrix into an element distribution matrix and noise intensity and improve the Sage filter to estimate the element distribution matrix. Second, we propose a calibration method to accurately diagnose the noise intensity deviation. We then propose a correct method to adaptively correct the noise intensity online. Third, under the assumption that the system is detectable, the unbiased and convergence of the proposed method is mathematically proven. Simulation results prove the effectiveness and reliability of the proposed method. Finally, we apply the proposed method to multiobject tracking of lidar and evaluate it on the official KITTI server. The proposed method on the KITTI pedestrian multiobject tracking leaderboard (http://www.cvlibs.net/datasets/kitti/eval_tracking.php) surpasses all existing methods using lidar, proving the feasibility of the method in practical applications. This work provides a new way to improve the performance of the Kalman filter and multiobject tracking.**

## I. INTRODUCTION

### A. Background

IN recent years, multiobject tracking, target behavior prediction, and improvement in driving safety of intelligent vehicles using lidar have been considered attractive research topics [1][2]. Adaptive filters are one of the most commonly used tracking methods [3], of which the Kalman filter (KF) [4] is a popular method.

KF is one of the most important developments of linear estimation theory; it is widely used and has significant theoretical interest. KF is the minimum mean square error estimator of the system state in a detectable system when the system model and noise statistics are known and satisfy the basic assumptions. However, when the noise covariance matrix is inaccurate, the filter will have problems with reduced accuracy or even divergence. The inaccuracy of the noise covariance matrix is mainly caused by the incorrect setting of the initial noise covariance matrix and the change of the noise covariance matrix over time [5]. Dunĭk et al. [5] studied the impact of incorrect noise covariance matrix settings on the filtering results and proved that incorrect specification of the measurement noise covariance matrix in the filter design might cause the filter accuracy to drop by nearly four orders of magnitude. To meet the conditions of the Kalman filter to achieve the best performance, different adaptive Kalman filter techniques have been proposed in recent years to estimate the noise covariance matrix. The principle of the adaptive Kalman filter is to dynamically adjust the filter's parameters by estimating unknown parameters such as the process and measurement noise covariance matrices from the measured data to achieve the best state estimation [6].

### B. Related work

Adaptive Kalman filtering methods are mainly divided into four groups: Bayesian methods, covariance matching methods, maximum-likelihood methods, and correlation methods.

The Bayesian method can be traced back to 1965 [7], mainly the state enhancement [8], multimodel, and variational Bayesian methods. The multimodel method includes the interacting multiple model method [9] and the particle method [10], which are sensitive to the prior selection of model number and parameters. The estimation accuracy of the variational Bayes method is not as good as that of the interactive multiple model method, but the computational efficiency is relatively improved [11]. Simo Särkkä et al. [12] applied the variational Bayesian method to the linear state-space model and estimated the process and measurement noise covariance matrices through the iteration of the Kalman filter. However, this method can only estimate the measurement noise covariance matrix.

The method of adopting a statistical framework and then selecting a model that maximizes the likelihood of the observed data as the estimation model is called the maximum likelihood method. This method was first proposed by Mehra [13] and later modified and improved by Maybeck [14], and then it has

This work was supported by National Key Research and Development Program of China (Nos. 2020AAA0108103, 2016YFD0701401, 2017YFD0700303 and 2018YFD0700602), Youth Innovation Promotion Association of the Chinese Academy of Sciences (Grant No. 2017488), Key Supported Project in the Thirteenth Five-year Plan of Hefei Institutes of Physical Science, Chinese Academy of Sciences (Grant No. KP-2019-16), Key Science and Technology Project of Anhui (Grant No. 202103a05020007) and Technological Innovation Project for New Energy and Intelligent Networked Automobile Industry of Anhui Province.

Chao Jiang and Shuhang Tan are with the Hefei Institutes of Physical Science, Chinese Academy of Sciences, Hefei 230031, China, and with the University of Science and Technology of China, Hefei 230026, China (e-mail: jc2009@mail.ustc.edu.cn; stan9177@mail.ustc.edu.cn). Zhiling Wang and Huawei Liang are with the Hefei Institutes of Physical Science, Chinese Academy of Sciences, Hefei 230031, China, also with the Anhui Engineering Laboratory for Intelligent Driving Technology and Application, Hefei 230031, China, and with the Innovation Research Institute of Robotics and Intelligent Manufacturing, Chinese Academy of Sciences, Hefei 230031, China (e-mail: zlwang@hfcas.ac.cn; hwliang@iim.ac.cn;).

been widely used [15]. However, the iterative computation of Maybeck [14] is extensive, and when the measurement vector dimension is smaller than the state vector dimension, the estimated process noise covariance matrix error is significant.

The covariance matching method is based on the standard linear or nonlinear state estimation algorithm, and it adjusts the noise covariance matrices according to the true system state and measurement estimation errors. For example, Sage et al. [16] proposed an adaptive filtering algorithm to estimate process and measurement noise covariance matrices online. However, this method uses some particular assumptions and approximations, so it cannot guarantee the convergence of noise covariance matrix estimates. Wang et al. [17] improved the Sage adaptive filter by using the property of innovation and applied it to power system state estimation. The disadvantage of this method is that the estimation performance is not stable, and the estimated noise covariance matrix may be nonpositive semidefinite.

The method that can estimate the noise covariance matrix by analysing the innovation sequence's autocorrelation of the asymptotically stable linear filter is called the correlation method. The correlation method includes the direct correlation method, indirect correlation method, weighted correlation method, and measurement average correlation method [5]. The performance of these methods is greatly affected by the user-defined filter gain K and initial parameters [18]. Dun k et al. [19] proposed a measurement difference correlation method (MDCM) based on the measurement average correlation method, which does not require the user to specify initial parameters such as gain K. This method has been proven to provide an unbiased estimation. However, when the measurement vector dimension is smaller than the state vector dimension, some elements of the process noise covariance matrix cannot be estimated. Rahul et al. [20] combined the advantages of Maybeck [14] and the measurement average correlation method and proposed another extension of the correlation method. This method has no strict restriction on the vector dimension. However, it is not suitable for unobservable systems, and the estimation performance is unstable when the process and measurement noise covariance matrices change non-stationary.

The main problems of the Bayesian method and maximum likelihood method are as follows: ① The process noise and measurement noise cannot be estimated simultaneously; ② when the dimension of the measurement vector is smaller than the dimension of the state vector, the error of the estimated process noise is significant. However, both process noise and measurement noise may be time-varying in practical applications. For example, in the multiobject tracking application of intelligent driving vehicles, lidar is used to track noncooperative targets. Because the detection accuracy of lidar is affected by distance, rain, and fog, the measurement noise is different at different distances and weather conditions. The process noise changes with the target's motion state; especially when the tracking object is a pedestrian with a changeable motion state, the process noise changes more frequently. In addition, in most target tracking tasks, the dimension of the measurement vector will be smaller than the dimension of the state vector. For example, in [21], the measurement vector includes the size and position, but the system state vector includes the size, position, and velocity.

Covariance matching and correlation methods do not have the above disadvantages, but their estimation performance is unstable. Chen et al. [22] used the prescribed boundary constraint method to enable the estimation error to converge. However, this method contains a series of complex design parameters, making the system design and implementation difficult. Gao et al. [23] used the chi-square test to diagnose whether the estimation of the noise covariance matrix was biased. However, they could not determine whether the bias came from the process or measurement noise covariance matrix. Sun et al. [24] monitor the estimator's performance by judging whether the autocorrelation of innovation approximately obeys a normal distribution. However, it only works when the process or measurement noise covariance matrices are known accurately.

*C. Motivation and contribution*

We found that the main reason these methods cannot correctly correct the estimation error when the estimated performance is unstable is that most studies do not provide an online calibration method that can simultaneously and accurately diagnose the deviation of the process and measurement noise covariance matrices. In response to these problems, this paper proposes a new adaptive filtering method with self-diagnosis and correction based on the covariance matching method to simultaneously estimate the process and measurement noise covariance matrices, and system state. We decompose the noise covariance matrix into an element distribution matrix and noise intensity. The noise intensity is defined as the sum of the elements of the noise covariance matrix. The element distribution matrix is defined as the noise covariance matrix divided by the noise intensity. The motivation for this is that we find that the estimation of noise covariance cannot converge to the true value, which is mainly caused by two reasons: ① the estimated noise covariance matrix is not positive semidefinite; ② the error between the estimated noise covariance modulus and the true noise covariance modulus is significant. Therefore, the main work of this paper is to propose a method that can ensure that the element distribution matrix is positive semidefinite while ensuring that the noise intensity converges to the true noise covariance modulus. An overview of our proposed noise covariance matrix estimation method for filters is shown in Figure 1. We summarize our contributions as follows:

- First, we improve the Sage filter [16] to adaptively estimate the element distribution matrix. The Sage filter is improved to avoid the estimated noise covariance matrix being nonpositive semidefinite.
- Second, we designed an autocovariance and Gaussian calibrator to diagnose the deviation between the noise intensity and the true value. We then proposed an online correction method to correct the noise intensity. Simulation experiments prove that the proposed method has good robustness in multiple types of systems and can improve the estimation accuracy of the system state.
- Third, under the assumption that the system is detectable (detectability is a condition to ensure the convergence of

the Kalman filter iteration), the unbiasedness and convergence of our proposed method are mathematically proven. (Process and measurement noise usually include multidimensional components. For example, the measurement noise of lidar includes three-dimensional size and position noise. If the element distribution matrix estimated by the improved Sage adaptive filter is biased, the distribution of noise intensity in each noise component will be inconsistent with the true value. The proof of convergence is to analyze whether the calibrator can accurately diagnose the deviation type and size and makes the diagnosis result converge to 0 when the noise intensity is close to the true value. Only when the calibrator converges can the correction method make the noise intensity converge to the true value.)

- Finally, we applied the proposed method to multiobject tracking of lidar and evaluated it on the official KITTI server. The proposed method on the KITTI pedestrian multiobject tracking leaderboard (http://www.cvlibs.net/datasets/kitti/eval_tracking.php) surpasses all methods using lidar, proving the feasibility of the method in practical applications.

The paper is organized as follows: Section II describes the problem statement and technical assumption. Section III presents the adaptive estimation method of the process and measurement noise covariance matrices and system state. Section IV analyses the estimation method's unbiasedness and convergence. Section V applies the proposed method to multiobject tracking. Section VI designs a series of simulation and multiobject tracking experiments. Finally, Section VII concludes the paper.

## II. PROBLEM STATEMENT

An LTV model of discrete time dynamic stochastic system

$$x_{k+1} = \Phi_k x_k + \varpi_k, \quad \varpi_k \sim N(0, Q_k), \tag{1}$$

$$z_k = H_k x_k + \upsilon_k, \quad \upsilon_k \sim N(0, R_k) \tag{2}$$

where the vectors $x_k \in \mathbb{R}^{n_x}$ and $z_k \in \mathbb{R}^{n_z}$ are the immeasurable system state and measurement at time instant k, respectively. $\Phi_k \in \mathbb{R}^{n_x \times n_x}$ and $H_k \in \mathbb{R}^{n_z \times n_x}$ are the state transition and observation matrix, respectively. $n_x$ is the state vector dimension, $n_z$ is the measurement vector dimension, and $n_z \leq n_x$. $\varpi_k \in \mathbb{R}^{n_x}$ and $\upsilon_k \in \mathbb{R}^{n_z}$ are the process and measurement noise, respectively. $Q_k$ and $R_k$ are the process and measurement noise covariance matrices, respectively.

The goal of adaptive noise covariance matrix estimation is to provide correct estimates of $Q_k$ and $R_k$ simultaneously and correct them online when they are biased.

### A. Problem Description

If the system parameters ($\Phi_k$, $H_k$, $Q_k$, and $R_k$) are known precisely, the baseline Kalman filter is the optimal estimator of the system given by (1). However, in most practical applications, $Q_k$ and $R_k$ are set incorrectly or change over time. Therefore, the problem that needs to be addressed in this paper can be stated: under the premise of satisfying the assumptions in Sec. II. B, given $\Phi_k$, $H_k$, and $z_k$, we formulate an adaptive algorithm to estimate system state $x_k$ and unknown process and measurement noise covariance matrices ($Q_k$ and $R_k$).

### B. Assumptions

The following assumptions are made:

*Assumption 2.1:* The state transition matrix $\Phi_k$ and the observation matrix $H_k$ are known, and the measurement sequence $z_k$ is accessible.

*Assumption 2.2:* $Q_k$ and $R_k$ are unknown and change slowly.

*Assumption 2.3:* The system is detectable; that is, any eigenvalue $\lambda_i$ of the state transition $\Phi_k$ satisfies $\text{Re}(\lambda_i) \geq 0$ and $\text{rank}([\Phi - \lambda_i I; H]) = n_x$, $i = 1, 2, ..., n_x$.

*Assumption 2.4:* $\varpi_k$ and $\upsilon_k$ are time-varying, mutually independent Gaussian white noises and independent of the initial state $x_0$. That is, $\mathrm{E}(\varpi_k) = \mathbf{0}^{n_x}$, $\mathrm{E}(\upsilon_k) = \mathbf{0}^{n_z}$, where $\mathrm{E}(\cdot)$ is the mathematical expectation, $\mathrm{cov}(\varpi_k, \varpi_l) = Q_k \sigma_{k,l} \in \mathbb{R}^{n_x \times n_x}$, $\mathrm{cov}(\upsilon_k, \upsilon_l) = R_k \sigma_{k,l} \in \mathbb{R}^{n_z \times n_z}$, where $\sigma_{k,l}$ is the Kronecker delta ($\sigma_{k,l} = 1$ if $k=l$, $\sigma_{k,l} = 0$ otherwise), and $\mathrm{cov}(\cdot)$ is the covariance.

## III. METHOD

An overview of the NC2 adaptive filter method is shown in Figure 1. The noise covariance matrix estimation is performed synchronously with the Kalman filter. Each filtering process performs a new estimation, calibration, and correction until the Gaussian calibration value converges to the threshold $T_G$. We named the proposed NC2 method with the initials of the three core technologies (normalized estimation, calibration, and correction).

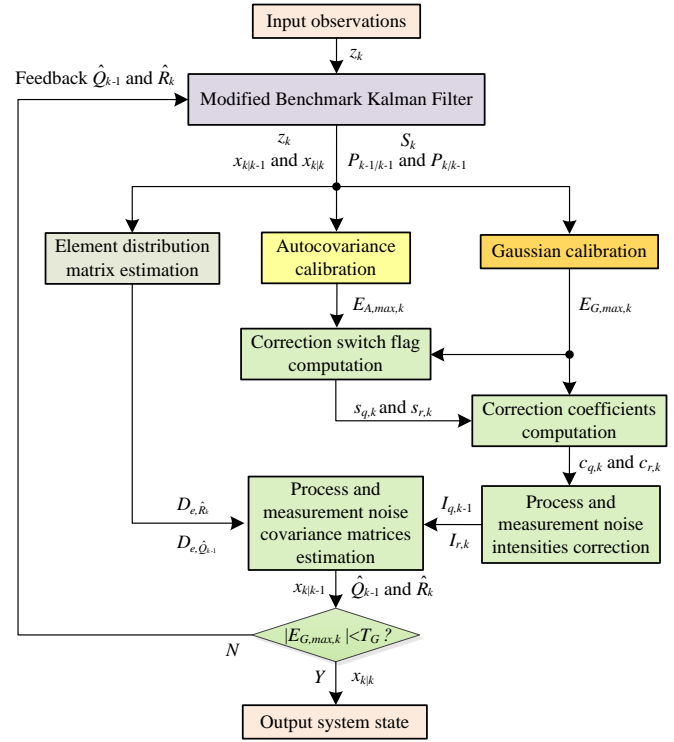

Fig. 1. Overview of our proposed NC2 noise covariance matrix estimation method for adaptive filters.

where $x_{k/k-1}$ and $x_{k/k}$ are the system state estimates of filter prediction and measurement update, respectively; $S_k$ is the innovation covariance; $P_{k/k-1}$ and $P_{k/k}$ are the state error covariance matrices of filter prediction and measurement update, respectively; $E_{A,max,k}$ and $E_{G,max,k}$ are autocovariance and Gaussian calibration values; $s_{q,k}$ and $s_{r,k}$ are calibration switch flags utilized to determine whether the corrector should correct the process or measurement noise; $D_{e,\hat{Q}_{k-1}}$ and $D_{e,\hat{R}_k}$ are the estimated element distribution matrices; $c_{q,k}$ and $c_{r,k}$ are correction coefficients; $I_{q,k-1}$ and $I_{r,k}$ are noise intensities; and $\hat{Q}_{k-1}$ and $\hat{R}_k$ are the estimated process and measurement noise covariance matrices, respectively.

### A. Modified Benchmark Kalman Filter

To achieve adaptive process and measurement noise covariance matrices and system state estimation, the benchmark Kalman filter is modified:

- Initialization

The state $x_0$ and the covariance matrix $P_0$ are initialized as an all-zero matrix, and the initialization of $Q_0$ and $R_0$ is defined in (39) in the experimental section.

- Prediction and Measurement update

$$\begin{cases} x_{k|k-1} = \Phi_{k-1} x_{k-1|k-1} \\ Q_{k-1|k-2} = \hat{Q}_{k-2} \\ P_{k|k-1} = \Phi_{k-1} P_{k-1|k-1} \Phi_{k-1}^T + Q_{k-1|k-2} \\ R_{k|k-1} = \hat{R}_{k-1} \\ S_k = H_k P_{k|k-1} H_k^T + R_{k|k-1} \\ K_k = P_{k|k-1} H_k^T S_k^{-1} \\ x_{k|k} = x_{k|k-1} + K_k (z_k - H_k x_{k|k-1}) \\ P_{k|k} = (I - K_k H_k) P_{k|k-1} (I - K_k H_k)^T + K_k R_{k|k-1} K_k^T \end{cases} \quad (3)$$

where $Q_{k-1|k-2}$ and $R_{k|k-1}$ are the process and measurement noise covariance matrices used in the filter, respectively, and $K_k$ is the Kalman gain. When $k=1$, $\hat{Q}_{k-2}$ and $\hat{R}_{k-1}$ are the initial noise covariance matrices. When $k>1$, $\hat{Q}_{k-2}$ and $\hat{R}_{k-1}$ are the estimates of the process and measurement noise covariance matrices at the previous time instant, respectively. We use the output of the filter at the present moment to estimate the process noise covariance $\hat{Q}_{k-1}$ and measurement noise covariance $\hat{R}_k$.

### B. Adaptive noise covariance matrix estimation

We decouple the estimation of $\hat{Q}_{k-1}$ and $\hat{R}_k$ into the estimation of the element distribution matrix ($D_{e,\hat{Q}_{k-1}}$ and $D_{e,\hat{R}_k}$) and noise intensity ($I_{q,k-1}$ and $I_{r,k}$), as shown in (4) and (5). The noise intensity is defined as the sum of the elements of the noise covariance matrix. The element distribution matrix is defined as the noise covariance matrix divided by the noise intensity.

$$\hat{Q}_{k-1} = I_{q,k-1} D_{e,\hat{Q}_{k-1}} \quad (4)$$

$$\hat{R}_k = I_{r,k} D_{e,\hat{R}_k} \quad (5)$$

Therefore, we need to propose a method to estimate the unbiased and positive semi-definite element distribution matrix and then propose a calibration and correction method to make the noise intensity converge to the true value.

*a). Element distribution matrix estimation*

Define the element distribution matrix as:

$$D_{e,\hat{R}_k} = \Lambda_{\hat{R}_k} / \sum_{i=1}^{n_z} \sum_{j=1}^{n_z} \Lambda_{\hat{R}_{k,i,j}} \quad (6)$$

$$D_{e,\hat{Q}_{k-1}} = \Lambda_{\hat{Q}_{k-1}} / \sum_{i=1}^{n_x} \sum_{j=1}^{n_x} \Lambda_{\hat{Q}_{k-1,i,j}} \quad (7)$$

where $\Lambda_{\hat{R}_k}$ is the estimation of the true measurement noise covariance matrix $R_k$; $\Lambda_{\hat{Q}_{k-1}}$ is the estimation of the true process noise covariance matrix $Q_{k-1}$; and $\Lambda_{\hat{R}_{k,i,j}}$ and $\Lambda_{\hat{Q}_{k-1,i,j}}$ represent the elements of the $i$th row and $j$th column in the $\Lambda_{\hat{R}_k}$ and $\Lambda_{\hat{Q}_{k-1}}$ matrices, respectively.

If $R_k$ and $Q_{k-1}$ do not change with time, we refer to the Sage filter [16] and define $\Lambda_{\hat{R}_k}$ and $\Lambda_{\hat{Q}_{k-1}}$ as:

$$\Lambda_{\hat{R}_k} = \frac{1}{N} \sum_{t=k-N+1}^{k} \tau_{I,t} \tau_{I,t}^T - H_k P_{k|k-1} H_k^T \quad (8)$$

$$\Lambda_{\hat{Q}_{k-1}} = H_k^{-1} (\frac{1}{N} \sum_{t=k-N+1}^{k} \tau_{I,t} \tau_{I,t}^T - \Lambda_{\hat{R}_k})(H_k^T)^{-1} - \Phi_{k-1} P_{k-1|k-1} \Phi_{k-1}^T \quad (9)$$

$\tau_{I,k}$ is innovation, defined as the difference between the filter's prediction and measurement.

$$\tau_{I,k} = z_k - H_k x_{k|k-1} \quad (10)$$

If $R_k$ and $Q_{k-1}$ change over time, exponential weights are used to emphasize the impact of recent data.

$$\Lambda_{\hat{R}_k} = \sum_{t=k-N+1}^{k} d_{N,b_1} b_1^{k-t} \tau_{I,t} \tau_{I,t}^T - H_k P_{k|k-1} H_k^T \quad (11)$$

$$\Lambda_{\hat{Q}_{k-1}} = H_k^{-1} (\sum_{t=k-N+1}^{k} d_{N,b_2} b_2^{k-t+1} \tau_{I,t} \tau_{I,t}^T - \Lambda_{\hat{R}_k}) / H_k^T - \Phi_{k-1} P_{k-1|k-1} \Phi_{k-1}^T \quad (12)$$

$$d_{N,b} = (1-b)/(1-b^N) \quad (13)$$

where $0<b<1$, $b_1=0.95$, $b_2=0.05$. $N$ is a constant when $k \leq 20$, $N = k-1$; when $k > 20$, $N = 20$.

In (12) and (13), let $\Lambda_{M_k} = \sum_{t=k-N+1}^{k} d_{N,b_2} b_2^{k-t+1} \tau_{I,t} \tau_{I,t}^T$ and compute $\Lambda_{\hat{R}_k}$ and $\Lambda_{\hat{Q}_{k-1}}$ by the recursive method.

$$\Lambda_{\hat{R}_k} = b_1 \Lambda_{\hat{R}_{k-1}} + d_{N,b_1} \tau_{I,k} \tau_{I,k}^T - d_{N,b_1} b_1^N \tau_{I,k-N} \tau_{I,k-N}^T \\ - H_k P_{k|k-1} H_k^T + b_1 H_{k-1} P_{k-1|k-2} H_{k-1}^T \quad (14)$$

$$\Lambda_{M_k} = b_2 \Lambda_{M_{k-1}} + d_{N,b_2} \tau_{I,k} \tau_{I,k}^T - d_{N,b_2} b_2^N \tau_{I,k-N} \tau_{I,k-N}^T \quad (15)$$

$$\Lambda_{\hat{Q}_{k-1}} = H_k^{-1} (\Lambda_{M_k} - \Lambda_{\hat{R}_k})(H_k^T)^{-1} - \Phi_{k-1} P_{k-1|k-1} \Phi_{k-1}^T \quad (16)$$

where, $\Lambda_{\hat{R}_0} = 0$, $\Lambda_{M_0} = 0$.

When $H_k$ is irreversible, we use the SVD method [25] to find its Moore-Penrose generalized inverse. When the generalized inverse matrix is not positive due to factors such as numerical calculation error, the $H_k^T$ approximation is used instead of the $H_k^{-1}$ approximation to keep the estimate of $Q_k$ semipositive.

$$\Lambda_{\hat{Q}_{k-1}} = H_k^T (\Lambda_{M_k} - \Lambda_{\hat{R}_k}) H_k - \Phi_{k-1} P_{k-1|k-1} \Phi_{k-1}^T \quad (17)$$

*b). Noise intensity estimation*

The noise intensity estimation and the element distribution estimation are performed independently. The initial noise intensity is set to a fixed value that is unrelated to the estimation of the element distribution matrix. To avoid the divergence of the estimation results of Eqs. (14) and (16), the initial noise intensity becomes extremely large; it is not recommended to initialize the noise intensity to $\sum_{i=1}^{n_x} \sum_{j=1}^{n_x} \Lambda_{\hat{Q}_{k-1,i,j}}$ and $\sum_{i=1}^{n_z} \sum_{j=1}^{n_z} \Lambda_{\hat{R}_{k,i,j}}$.

As shown in Figure 1, in order to be able to diagnose the deviation of the process and measurement noise intensity at the same time, we propose a dual calibrator joint calibration method. The noise intensity can be gradually corrected to the true value through the iterative calculation of the calibrator and the corrector. The noise intensity when the calibration value is close to 0 is regarded as the true noise intensity.

*1). Autocovariance Calibration*

Autocovariance calibrator is defined as:

$$\begin{cases} E_{A,k} = \tau_{A,k}([\frac{N}{3}:l_{step}:N]) \\ \tau_{A,k}(\eta) = \frac{1}{\eta - i} \sum_{t=k-\eta+i}^{k}(\tau_{I,k,t} - \bar{\tau}_{I,k})(\tau_{I,k,t-i} - \bar{\tau}_{I,k})^T \\ \eta \in [\frac{N}{3}:5:N] \end{cases} \quad (18)$$

where $E_{A,k}$ is the autocovariance calibration matrix; $l_{step}$ is the step size; $\bar{\tau}_{I,k}$ is the sample mean of the innovation sequence, and $\bar{\tau}_{I,k} = \sum_{t=k-N}^{k} \tau_{I,k,t}/N$; $i$ is defined as the translation step length of the innovation sequence relative to itself; and $i = 1$.

*2). Gaussian Calibration*

Gaussian calibrator is defined as:

$$E_{G,k} = (P_{e,k} - P_{g,k})/\sqrt{P_{e,k}^2 + P_{g,k}^2} \quad (19)$$

where $E_{G,k}$ is the Gaussian calibration matrix and $P_{g,k}$ is defined as the ratio of the absolute value sequence of innovation's expected $\mu_{|\tau|}$ to the theoretical confidence interval threshold $T_{g,k}$ of the Gaussian distribution. $P_{e,k}$ is defined as the ratio of the innovation absolute value sequence's sample mean $\hat{\mu}_{|\tau|}$ to the estimated value $\hat{T}_{g,k}$ of the theoretical threshold $T_{g,k}$.

$$P_{g,k} = \mu_{|\tau|}/T_{g,k} \quad (20)$$

$$P_{e,k} = \hat{\mu}_{|\tau|}/\hat{T}_{g,k} \quad (21)$$

where:

$$\mu_{|\tau|} = E(|\tau_{I,k}|) = \int_{-\infty}^{+\infty} \frac{|\tau_{I,k}|e^{-\tau_{I,k}^2/2S_{t,k}}}{\sqrt{2\pi S_{t,k}}} d\tau_{I,k} = \frac{2\sqrt{S_{t,k}}}{\sqrt{2\pi}} \quad (22)$$

$$\hat{\mu}_{|\tau|} = \sum_{t_k=k-N+1}^{k}|\tau_{I,k}|/N \quad (23)$$

where $S_{t,k}$ is the theoretical innovation covariance and $T_{g,k}$ is the theoretical confidence interval threshold of the Gaussian distribution. According to the properties of the Gaussian distribution, when the confidence is 95%,

$$T_{g,k} = 2\sqrt{S_{t,k}}, \quad \hat{T}_{g,k} = 2\sqrt{S_k} \quad (24)$$

Therefore,

$$P_{g,k} = \mu_{|\tau|}/T_{g,k} = 1/\sqrt{2\pi}$$

$$P_{e,k} = \frac{\sum_{t_k=k-N+1}^{k}|\tau_{I,k}|}{2N\sqrt{S_k}} \quad (25)$$

The Gaussian calibration value $E_{G,k}$ can be obtained by substituting $P_{e,k}$ and $P_{g,k}$ into (19).

*3). Noise intensity estimation*

We use statistical methods to construct the calibrator. However, due to the limited sample size of the innovation sequence, there may be minor deviations between the diagnostic results of the calibrator and the true value. To improve the calibration sensitivity, we use (19) to take the element with the largest absolute value in the matrices $E_{A,k}$ and $E_{G,k}$ as the autocovariance calibration value and Gaussian calibration value, respectively.

$$E_{c,\max,k} = \begin{cases} \max(E_{c,k}) & |\max(E_{c,k})| > |\min(E_{c,k})| \\ \min(E_{c,k}) & \text{else} \end{cases}, \quad c \in \{A,G\} \quad (26)$$

where $\max(\cdot)$ and $\min(\cdot)$ are functions to find the matrix's maximum and minimum element values, respectively.

*Theorem 2:* The filter described by (3) is used in the discrete-time dynamic random system described by (1) and (2). Under Assumptions 2.1, 2.2, 2.3 and 2.4, the autocovariance calibration values $E_{A,\max,k}$ and Gaussian calibration values $E_{G,\max,k}$ are computed by the calibrators described by (18), (19) and (26) and have the following properties:

*Property 1:* $E_{A,\max,k}$ converges to 0 when the estimation of the noise covariance matrix is equal to the true noise covariance matrix or the estimated deviation of the process noise covariance matrix and the measurement noise covariance matrix is equal (i.e., $\delta_{\varpi,k} = \delta_{v,k} = 1$).

*Property 2:* $E_{A,\max,k}$ is greater than 0 when $\delta_{\varpi,k} > 1 \& \delta_{v,k} \leq 1$ or $\delta_{\varpi,k} \geq 1 \& \delta_{v,k} < 1$, and $E_{A,\max,k}$ is less than 0 when $\delta_{\varpi,k} < 1 \& \delta_{v,k} \geq 1$ or $\delta_{\varpi,k} \leq 1 \& \delta_{v,k} > 1$.

*Property 3:* $E_{G,\max,k}$ converges to 0 when the estimate of the noise covariance matrix is equal to the true noise covariance matrix (i.e., $\delta_{\varpi,k} = \delta_{v,k} = 1$).

*Property 4:* $E_{G,\max,k}$ is greater than 0 when $\delta_{\varpi,k} < 1 \& \delta_{v,k} \leq 1$ or $\delta_{\varpi,k} \leq 1 \& \delta_{v,k} < 1$, and $E_{G,\max,k}$ is less than 0 when $\delta_{\varpi,k} > 1 \& \delta_{v,k} \geq 1$ or $\delta_{\varpi,k} \geq 1 \& \delta_{v,k} > 1$. $E_{G,\max,k}$ is a monotonically decreasing function of $\delta_{\varpi,k}$ and $\delta_{v,k}$, and $E_{G,\max,k} < 1$.

where $\delta_{v,k} = I_{q,k}/\sum_{i=1}^{n_z}\sum_{j=1}^{n_z}Q_{k,ij}$, $\delta_{v,k} \in \mathbb{R}+$; $\delta_{\varpi,k} = I_{r,k}/\sum_{i=1}^{n_z}\sum_{j=1}^{n_z}R_{k,ij}$, $\delta_{\varpi,k} \in \mathbb{R}+$.

proof: see Apendix B.

According to Theorem 1, the sign of $E_{A,\max,k}$ and $E_{G,\max,k}$ is the same when there is a deviation in the process noise, and the sign of $E_{A,\max,k}$ and $E_{G,\max,k}$ is the opposite when there is a deviation in the measurement noise. Therefore, we set two calibration switch flag $s_{q,k}$, $s_{r,k}$.

if $(E_{A,\max,k} * E_{G,\max,k} > 0)$, $s_{q,k}=1$, $s_{r,k}=0$

if $(E_{A,\max,k} * E_{G,\max,k} < 0)$, $s_{q,k}=0$, $s_{r,k}=1$

if $(E_{A,\max,k} * E_{G,\max,k} = 0)$, $s_{q,k}=0$, $s_{r,k}=0$ (27)

That is, the process noise covariance matrix is corrected when $E_{A,\max,k}$ and $E_{G,\max,k}$ have the same sign. The measurement noise covariance matrix is corrected when $E_{A,\max,k}$ and $E_{G,\max,k}$ have different signs.

According to Property 4 of Theorem 1, we design negative feedback correction coefficients $c_{q,k}$ and $c_{r,k}$ to correct the noise intensity.

$$c_{i,k}=1+\sigma * s_{i,k} * E_{G,max,k} \quad (i \in \{q,r\}) \tag{28}$$

$$I_{q,k-1}= I_{q,k-2} * c_{q,k-1}, \ I_{r,k}= I_{r,k-1} * c_{r,k} \tag{29}$$

where $\sigma$ is the adjustment coefficient of the convergence rate, $1<\sigma<1$. When $E_{G,max,k}<0$, the noise intensity will be amplified. Similarly, when $E_{G,max,k}>0$, the noise intensity will be reduced.

In practical applications, when the estimate of the noise covariance matrix is close to the true noise covariance matrix, due to the statistical error, the Gaussian calibration value $E_{G,max,k}$ oscillates around 0 instead of being fixed to 0. To avoid unnecessary correction, we set a correction threshold $T_G$ ($T_G$ =0.02). When $|E_{G,max,k}|<T_G$, set $E_{G,max,k}$ =0. According to the definitions of Equation (27) and (28), at this moment, the correction coefficients $c_{q,k}$ and $c_{q,k}$ are both equal to 1 and the noise intensity remains unchanged; that is, the correction process stops.

## IV. UNBIASEDNESS AND CONVERGENCE ANALYSIS

To judge whether the process and measurement noise covariance matrices can be correctly estimated by using (4) and (5), it is necessary to judge whether the estimated element distribution matrices $D_{e,\hat{Q}_{k-1}}$ and $D_{e,\hat{R}_k}$ are unbiased and whether the estimate of noise intensity is convergent.

### A. Unbiased analysis of element distribution estimation

*Theorem 1:* the filter described by equation (3) in the discrete-time dynamic stochastic system described by equations (1) and (2), under Assumptions 2.1, 2.2, 2.3, and 2.4, the element distribution matrices $D_{e,\hat{Q}_{k-1}}$ and $D_{e,\hat{R}_k}$ described by equations (8) and (9) are unbiased estimates of the normalized process and the measurement noise covariance matrices, respectively.

*Proof:* In the Kalman filter, the process and measurement noise covariance matrices can be described as:

$$\begin{aligned}Q_{k-1} &= P_{k|k-1} - \Phi_{k-1}P_{k-1|k-1}\Phi_{k-1}^T \\ &= H_k^{-1}(E(\tau_{I,k}\tau_{I,k}^T) - R_k)(H_k^T)^{-1} - \Phi_{k-1}P_{k-1|k-1}\Phi_{k-1}^T\end{aligned} \tag{30}$$

$$R_k = E(\tau_{I,t}\tau_{I,k}^T) - H_k P_{k|k-1}H_k^T \tag{31}$$

The expectation of $D_{e,\hat{R}_k}$ is:

$$\begin{aligned}E(\Lambda_{\hat{R}_k}) &= E(\sum_{t=k-N+1}^{k} d_{N,b_1} b_1^{k-t} \tau_{I,t}\tau_{I,t}^T - H_k P_{k|k-1}H_k^T) \\ &= \sum_{t=k-N+1}^{k} d_{N,b_1} b_1^{k-t} E(\tau_{I,t}\tau_{I,t}^T) - H_k P_{k|k-1}H_k^T \\ &= (1-b_1)/(1-b_1^N)\sum_{t=0}^{N-1} b_1^t E(\tau_{I,t}\tau_{I,t}^T) - H_k P_{k|k-1}H_k^T \\ &= E(\tau_{I,t}\tau_{I,t}^T) - H_k P_{k|k-1}H_k^T = R_k\end{aligned} \tag{32}$$

$$E(D_{e,\hat{R}_k}) = E(\Lambda_{\hat{R}_k})/\sum_{i=1}^{m}\sum_{j=1}^{n} E(\Lambda_{\hat{R}_{k,i,j}}) = R_k/\sum_{i=1}^{m}\sum_{j=1}^{n} R_{k,ij} \tag{33}$$

The expectation of $D_{e,\hat{Q}_{k-1}}$ is:

$$\begin{aligned}&E(\Lambda_{\hat{Q}_{k-1}}) \\ &= H_k^{-1}(\sum_{t=k-N+1}^{k} d_{N,b_2} b_2^{k-t+1} E(\tau_{I,t}\tau_{I,t}^T) - E(\Lambda_{\hat{R}_k}))/H_k^T - \Phi_{k-1}P_{k-1|k-1}\Phi_{k-1}^T \\ &= H_k^{-1}(E(\tau_{I,t}\tau_{I,t}^T) - E(\Lambda_{\hat{R}_k}))(H_k^T)^{-1} - \Phi_{k-1}P_{k-1|k-1}\Phi_{k-1}^T = Q_{k-1}\end{aligned} \tag{34}$$

$$E(D_{e,\hat{Q}_{k-1}}) = E(\Lambda_{\hat{Q}_{k-1}})/\sum_{i=1}^{n}\sum_{j=1}^{n} E(\Lambda_{\hat{Q}_{k-1,i,j}}) = Q_{k-1}/\sum_{i=1}^{m}\sum_{j=1}^{n} Q_{k-1,ij} \tag{35}$$

Therefore, it can be proven that $D_{e,\hat{Q}_{k-1}}$ and $D_{e,\hat{R}_k}$ are unbiased estimates of the normalized process and measurement noise covariance matrices.

Prove up.

### B. Convergence analysis of noise intensity estimation

When the process noise intensity is underestimated, that is, $\delta_{b,k}<1$, $I_{q,k} < \sum_{i=1}^{n_x}\sum_{j=1}^{n_x} Q_{k,ij}$. According to property 2 and property 4 of Theorem 1, both $E_{A,max,k}$ and $E_{G,max,k}$ are greater than 0.

According to equations (27) and (28), we can obtain:

$$-c_{r,k}=1, \ c_{q,k}>1.$$

Let $\sigma$ be small enough that $c_{q,k}*\delta_{b,k} \le 1$, then:

$$\delta_{q,k+n+1} = \delta_{q,k+n}c_{q,k+n} > \delta_{q,k+1} > \delta_{q,k} \tag{36}$$

Therefore, $\{\delta_{q,k}\}$ is a monotonically increasing sequence, and $\delta_{q,k} \le 1/c_{q,k} \le 1$; according to the famous squeeze theorem[26], we can obtain:

$$\forall \varepsilon > 0, \ \exists N \in N^+, \ \text{when} \ n>N, |\delta_{q,k+n}-1|<\varepsilon, \ |I_{q,k+n} - \sum_{i=1}^{n_x}\sum_{j=1}^{n_x} Q_{k,ij}|<\varepsilon.$$

That is, $I_{q,k}$ converges to $\sum_{i=1}^{n_x}\sum_{j=1}^{n_x} Q_{k,ij}$. Similarly, it can be proven that $I_{q,k}$ converges to $\sum_{i=1}^{n_x}\sum_{j=1}^{n_x} Q_{k,ij}$ when $I_{q,1} > \sum_{i=1}^{n_x}\sum_{j=1}^{n_x} Q_{1,ij}$, and $I_{r,k}$ converges to $\sum_{i=1}^{n_z}\sum_{j=1}^{n_z} R_{k,ij}$ when $I_{r,k} \ne \sum_{i=1}^{n_z}\sum_{j=1}^{n_z} R_{k,ij}$.

Prove up.

Therefore, through the iterative calculation of the calibrator and the corrector, the estimation of the noise intensity can converge to the sum of the elements of the true noise covariance matrix.

## V. APPLICATION TO MULTIOBJECT TRACKING

Multi-object tracking (MOT) is essential for many applications, such as intelligent vehicles and flying robots [27]. In the practical application of multi-object tracking, the measurement noise is time-varying due to the impact of the environment on the sensor. In addition, the frequent changes in the motion state of the tracked object make the process noise also change with time. Therefore, it is necessary to adopt the NC2 adaptive method in the tracking system.

The overall architecture of the MOT system is mainly divided into the following steps: ① Using the lidar point cloud detector or clustering algorithm to obtain the object's 3D detection results $z_k$. ② Using the NC2 filter to predict the tracked object state $x_{k|k-1}$. ③ Using Euclidean distance and the Hungarian algorithm to match the detection $z_k$ with the prediction state $x_{k|k-1}$. ④ In the NC2 filter, the predicted state $x_{k|k-1}$ is updated based on its matched detection $z_k$. ⑤ Unmatched detection will be used to generate a new trajectory $T_{new}$, and the trajectory that does not match any detection will be deleted.

### A. Object Detection

The existing point cloud detection methods mainly include deep learning and 3D clustering methods. The generalization ability of deep learning methods is poor, but their detection accuracy is usually higher than that of traditional clustering methods. To facilitate the analysis of the performance of the NC2 tracking system when using different precision detection results, we use the 3D clustering method [28] in the KITTI MOT validation dataset [29] and the deep learning method (Voxel R-CNN [30]) in the KITTI MOT test dataset [29]. Both the clustering method and Voxel R-CNN can output a series of measurements $D_k = \{z_{1,k}, z_{2,k}, ..., z_{n_t,k}\}$, where $n_t$ is the number of objects detected. The object's measurement vector obtained by the 3D clustering or Voxel R-CNN method is $z_{i,k} = [p_{x,k} \ p_{y,k} \ p_{z,k} \ l_k \ w_k \ h_k]$. where $(p_{x,k}, p_{y,k}, p_{z,k})$ and $(l_k, w_k, h_k)$ are the 3D position and size of the object.

### B. NC2 Filter: State Prediction

In the discrete-time dynamic system described in (1) and (2), we use the first-order constant velocity model to describe the motion mode of the tracked object and use the NC2 adaptive method to estimate the noise covariance matrix and system state of the tracked object. Let:

$$\Phi_k = \begin{bmatrix} 1 & 0 & 0 & 0 & 0 & 0 & 1 & 0 & 0 \\ 0 & 1 & 0 & 0 & 0 & 0 & 0 & 1 & 0 \\ 0 & 0 & 1 & 0 & 0 & 0 & 0 & 0 & 1 \\ 0 & 0 & 0 & 1 & 0 & 0 & 0 & 0 & 0 \\ 0 & 0 & 0 & 0 & 1 & 0 & 0 & 0 & 0 \\ 0 & 0 & 0 & 0 & 0 & 1 & 0 & 0 & 0 \\ 0 & 0 & 0 & 0 & 0 & 0 & 1 & 0 & 0 \\ 0 & 0 & 0 & 0 & 0 & 0 & 0 & 1 & 0 \\ 0 & 0 & 0 & 0 & 0 & 0 & 0 & 0 & 1 \end{bmatrix}, \quad H_k = \begin{bmatrix} 1 & 0 & 0 & 0 & 0 & 0 & 0 & 0 & 0 \\ 0 & 1 & 0 & 0 & 0 & 0 & 0 & 0 & 0 \\ 0 & 0 & 1 & 0 & 0 & 0 & 0 & 0 & 0 \\ 0 & 0 & 0 & 1 & 0 & 0 & 0 & 0 & 0 \\ 0 & 0 & 0 & 0 & 1 & 0 & 0 & 0 & 0 \\ 0 & 0 & 0 & 0 & 0 & 1 & 0 & 0 & 0 \end{bmatrix} \quad (37)$$

Define the system state as $T_{s,k} = \{x_{1,k}, x_{2,k}, ..., x_{n_m,k}\}$, $x_{j,k} = [p_{x,k} \ p_{y,k} \ p_{z,k} \ l_k \ w_k \ h_k \ v_{x,k} \ v_{y,k} \ v_{z,k}]$. where each state vector $x_{j,k}$ represents a tracking trajectory, $n_m$ represents the number of tracking trajectories, and $(v_{x,k}, v_{y,k}, v_{z,k})$ represents the 3D velocity of the object. Both the process and measurement noise covariance matrices are initialized to a matrix of all-ones. The predicted system state is $T_{est,k}$, $T_{est,k} = \Phi_k T_{s,k} = \{x_{1,k|k-1}, x_{2,k|k-1}, ..., x_{n_m,k|k-1}\}$.

### C. Object Matching

To improve matching efficiency, we designed a cascade matching scheme. First, calculate the Euclidean distance $d_{eu,k}^i$ between the detection $D_k$ and the prediction $T_{est,k}$.

$$\begin{cases} d_{eu,k}^{i,j} = \sqrt{\sum_{t=1}^{2}(z_{i,k}[t] - x_{j,k|k-1}[t])^2} \\ T_{md} = \sqrt{\sum_{t=1}^{2}\hat{R}_{k,t,t}}/2 \end{cases} \quad (38)$$

If $d_{eu,k}^i < T_{md}$, consider that $z_{i,k}$ may be the corresponding detection of $x_{j,k|k-1}$, further compute the 3D Intersection of Union (IoU) between $z_{i,k}$ and detection $x_{j,k|k-1}$, and store the calculation result in a similarity matrix with dimensions of $n_t * n_m$. Otherwise, stop further matching, and set the element value of the corresponding coordinate of the similarity matrix to 0. Finally, the Hungarian algorithm is used to solve the optimal matching pair in the similarity matrix.

The output of the matching is a series of matching detection $D_{match,k}$ and corresponding prediction state $T_{match,k}$, and unsuccessful matching detection $D_{unmatch,k}$ and prediction state $T_{unmatch,k}$.

### D. NC2 Filtering: Noise Covariance and State Update

We use the method described in equation (3) to update the predicted state and corresponding state covariance and use equations (4) and (5) to update the process and measurement the noise covariance matrices. The update of each trajectory is performed independently, and the predicted state $T_{match,k}^j$ of each trajectory is updated using the matching measurement vector $D_{match,k}^j$.

### E. Life Cycle Management

Since all unmatched detections may be new objects, creating new tracking trajectories for unmatched detections is necessary. However, a new tracking trajectory $T_{s,k}^j$ is generated only when the object is continuously detected for more than two frames to avoid tracking false positive objects. When the j-*th* object disappears from the scene in the k+n *th* frame, the predicted state of its tracking trajectory $T_{est,k+n}^j$ cannot match any detection. To avoid false tracking, the tracking trajectory of the disappearing object needs to be deleted. However, to avoid matching failures caused by occlusion or missed detection, the tracker is set to use the kinematics model to keep tracking $T_{est,k+n}^j$ for two frames after the matching fails. If and only $T_{est,k+n+1}^j$ and $T_{est,k+n+2}^j$ fail to match any detection results, the tracking trajectory is deleted.

## VI. EXPERIMENTS

The experiment includes four parts: calibrator evaluation, performance evaluation in time-varying systems, performance evaluation in multi-class systems, and the application of the NC2 adaptive filter in LIDAR point cloud multiobject tracking.

### A. Calibrator evaluation

Let $\delta_\varpi \in [0.1, 1, 10]$, $\delta_\upsilon \in [0.1, 1, 10]$, compute the Gaussian and autocovariance calibration values in the system defined in Ref.[20], and the results are shown in Figure 2.

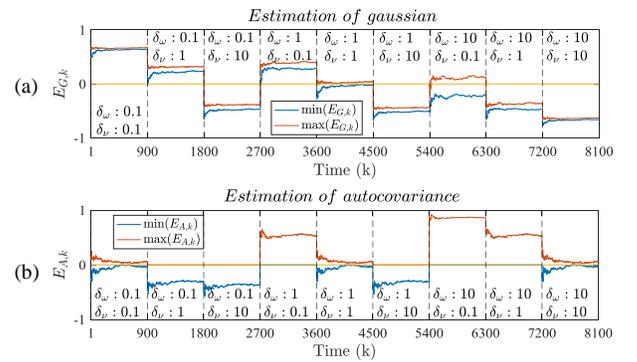

Fig. 2. Relationship between calibration value and estimated deviation (a), Gaussian calibration value (b), Autocovariance calibration value.

In Figure 2, $\delta_\varpi$ and $\delta_\upsilon$ are changed every 900 frames. It is obvious that the results in Figure 2 are consistent with the conclusion of Theorem 1. We noticed that when $\delta_\varpi = \delta_\upsilon = 0.1$ and $\delta_\varpi = \delta_\upsilon = 10$, the autocovariance calibration value oscillates around 0. This case does not affect the normal correction because the autocovariance calibration value is only used to determine whether the corrector should first correct the process noise or the measurement noise. When the filter has the same noise estimation deviation, it is reasonable first to correct the process noise or measurement noise.

*B. Estimation performance in time-varying noise systems*

As this paper studies the noise time-varying system, we multiply the time-invariant noise covariance matrix in reference [20] by the time-varying coefficients to obtain the time-varying noise covariance matrix. The F norm of the true noise covariance matrix ($Q_{true,k}$ and $R_{true,k}$) is shown as the blue line in Figure 3. The initial noise covariance matrix ($\hat{Q}_0$ and $\hat{R}_0$) of the estimator is randomly generated by (39), and the F norm of the initial noise covariance matrix is marked with the black triangle in Figure 3. The orange line represents the F norm of the corrected noise covariance matrix ($\hat{Q}_k$ and $\hat{R}_k$) in Figure 3.

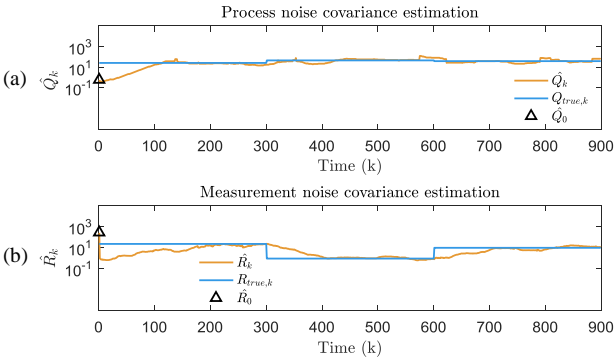

Fig. 3. The correction result of noise covariance matrices in noise time-varying system (a), process noise covariance matrix (b), measurement noise covariance matrix.

Figure 3 proves that when the noise covariance in the filter is set incorrectly and time-varying, our proposed method can correct the noise covariance matrix online and enable the correction result to converge to the ground truth.

*C. Estimation performance in multi-class systems*

In this subsection, we evaluate the performance of the noise covariance matrix estimation method in multiple classes of systems. Related research usually conducts evaluation experiments in a given system with specified parameters. Strictly speaking, its conclusions only apply to the systems and parameters specified in the experiment. It is unreasonable to extend its conclusions to other parameters or even other classes of systems. Our evaluation method is as follows: First, L repeated random experiments are used to evaluate the average estimation performance of the adaptive Kalman filtering method in the specified category system. In each repeated experiment, the system parameters are randomly generated by Algorithm 1 (see Appendix C), and the measurement sequence with time-varying noise is generated by Algorithm 2 (see Appendix C). Second, we repeat the above experiments in multiple classes of systems to evaluate the applicability of the noise covariance matrix estimation method in each class of systems.

Related research is usually evaluated in observable systems. However, the convergence condition of the Kalman filter is that the system is detectable, so it is necessary to evaluate both the observable system (OB) and detectable but unobservable system (UO). In the introduction section, we mentioned that when the measurement vector's dimension is less than the state vector's dimension, the performance of the noise covariance estimation method may decrease. Therefore, the system with the measurement vector's dimension less than the state vector's dimension needs to be analysed separately. In this paper, we consider the following systems:
- Observable system ($n_x=3$, $n_z=2$);
- Observable system ($n_x=n_z=3$);
- Detectable but unobservable system (including $n_x=n_z=3$ and $n_x=3$, $n_z=2$).

Since the detectable but unobservable system is not the focus of this paper, we do not separately analyse the case where the measurement vector's dimension is less than the state vector's dimension in this class of system. In the three types of systems mentioned above, we compare our NC2 estimation method with the advanced noise estimation methods that provide open source codes, including Maybeck [14][1], MDCM [19][1], Rahul [20][2], ROSE [31][3], and Sage [16]. Sage has not provided open-source code, so we reproduced the improved version of Sage [17].

*a). Experimental data generation method*

For the method of generating system parameters and measurement series, please refer to Appendix C.

Verifying whether the estimation algorithm can estimate the correct noise covariance matrix from the filter with the wrong initial noise covariance matrix is one of the purposes of the experiment. Therefore, we initialize the noise covariance matrix using the following randomly generated process and the measurement noise covariance matrices.

$$\begin{cases} \hat{Q}_0 = Q \times c_5^{(\text{randn}(n_x)+0.5)} \\ \hat{R}_0 = R \times c_5^{(\text{randn}(n_z)+0.5)} \end{cases} \quad (39)$$

where $c_5=10$ and $Q$ and $R$ are the true noise covariance matrices generated by Algorithm 1.

*b). Experimental results*

The experimental evaluation metrics are the average estimation error of the process and measurement noise covariance matrices, divergence rate (DR), inability rate (IR), and system state estimation error of the Kalman filter. The divergence rate refers to the probability that the estimated noise covariance matrix cannot converge. The inability rate refers to the probability that the code cannot be executed. If the estimation result of the noise covariance estimation algorithm diverges or fails, it means that the algorithm is not suitable for the tested system parameter. For each type of system studied in this paper, the average estimation error and standard deviation are used to evaluate the estimation accuracy of the algorithm for this kind of system; the average divergence rate and inability rate are used to evaluate the applicability and robustness of the estimation algorithm to this kind of system.

*1). The average estimation error and standard deviation of L*

[1] https://github.com/USNavalResearchLaboratory/TrackerComponentLibrary
[2] https://github.com/BenGravell/adaptive-kalman-filter
[3] https://github.com/konimarti/kalman

*experiments are defined as:*

$$\Delta c_j = \sum_{k=1}^{T_m} \|\hat{c}_{j,k} - c_{true,j,k}\|_F / (T_m \|c_{true,j,k}\|_F), \quad c \in \{Q, R\} \quad (40)$$

$$\overline{\Delta c} = \log\left(\sum_{j=1}^{L} \Delta c_j / L\right), \quad c \in \{Q, R\} \quad (41)$$

$$\sigma_{\Delta c} = \log\left(\sqrt{\sum_{j=1}^{L}(\Delta c_j - \overline{\Delta c})^2 / (L-1)}\right), \quad c \in \{Q, R\} \quad (42)$$

where $j \in \{1, 2, \ldots, L\}$; $Q_{true,j,k}$ and $R_{true,j,k}$ are generated in Algorithm 2, which are the true noise covariance matrices at time instant $k$ of the $j$th experiment; and $\hat{Q}_{j,k}$ and $\hat{R}_{j,k}$ are the estimations of the process and measurement noise covariance matrices at time instant $k$ of the $j$th experiment. $\Delta Q_j$ and $\Delta R_j$ are the estimation errors of the process and measurement noise covariance matrices at the $j$th experiment; $\overline{\Delta Q}$ and $\overline{\Delta R}$ are the average estimation errors of the process and measurement noise covariance matrices, respectively; $\sigma_{\Delta Q}$ and $\sigma_{\Delta R}$ are the standard deviation of the estimation error of the process and measurement noise covariance, respectively.

Table I illustrates various adaptive Kalman filtering methods' average estimation error and standard deviation in the three types of systems mentioned above. The symbol "nan" in the table indicates that the code of the adaptive Kalman filtering method cannot be executed correctly in the corresponding system. Obviously, the noise covariance matrix estimated by our proposed method (NC2) in all three mentioned systems has a minor estimation error and standard deviation. Table II shows that the main reason for the poor performance of other methods is the divergence or inability of the filter under some system parameters.

THE AVERAGE ESTIMATION ERROR AND STANDARD DEVIATION OF THE NOISE COVARIANCE MATRIX ESTIMATION METHOD IN DIFFERENT SYSTEMS.

| Method | OB ($n_x=n_z$) | | | | OB ($n_x>n_z$) | | | | UO | | | |
|---|---|---|---|---|---|---|---|---|---|---|---|---|
| | $\overline{\Delta Q}$ | $\overline{\Delta R}$ | $\sigma_{\Delta Q}$ | $\sigma_{\Delta R}$ | $\overline{\Delta Q}$ | $\overline{\Delta R}$ | $\sigma_{\Delta Q}$ | $\sigma_{\Delta R}$ | $\overline{\Delta Q}$ | $\overline{\Delta R}$ | $\sigma_{\Delta Q}$ | $\sigma_{\Delta R}$ |
| Uncorrect | 5.36 | 5.15 | 6.76 | 6.15 | 5.37 | 4.39 | 6.69 | 5.35 | 5.27 | 5.32 | 6.75 | 6.32 |
| Rahul | 5.00 | 2.54 | 6.65 | 5.42 | 5.37 | 0.29 | 6.69 | 0.78 | nan | nan | nan | nan |
| Sage | 4.07 | 0.25 | 5.75 | 0.11 | 5.89 | 1.12 | 8.50 | 4.07 | 37.7 | 3.97 | 40.8 | 5.67 |
| ROSE | 47.7 | 0.56 | 51.1 | 0.29 | 5.36 | 0.45 | 8.66 | 0.21 | 3.17 | **0.30** | 6.35 | **-0.01** |
| Maybeck | 9.48 | 4.87 | 11.9 | 6.52 | 16.5 | 4.91 | 19.7 | 6.66 | 0.48 | 2.32 | 0.21 | 5.38 |
| MDCM | 13.3 | 4.66 | 16.6 | 6.53 | 26.7 | 4.35 | 28.1 | 5.41 | 22.1 | 22.2 | 25.5 | 25.6 |
| Our | **0.61** | **0.01** | **0.39** | **-0.17** | **0.59** | **-0.28** | **0.41** | **-0.51** | **0.36** | 0.34 | **0.16** | 0.07 |

*2). The divergence rate is defined as:*

$$P_{d,c} = \sum_{j=1}^{L}(\text{sgn}(\Delta c_j - 15) + 1)/2L, \quad c \in \{Q, R\} \quad (43)$$

where $\text{sgn}(\cdot)$ is a symbolic function and $P_{d,Q}$ and $P_{d,R}$ are the divergence rate of the process and measurement noise covariance, respectively. When the code of the noise covariance estimation method is executed incorrectly, let $\Delta Q_j = \Delta R_j = T_e$ ($T_e$ is set to a very large value, let $T_e = 10^4$). The inability rate is expressed as:

$$P_{I,c} = 1 - \sum_{j=1}^{L}(\text{sgn}(\Delta c_j - T_e) + 1)/2L, \quad c \in \{Q, R\} \quad (44)$$

-where $P_{I,Q}$ and $P_{I,R}$ are the inability rate of the process and measurement noise covariance, respectively.

In different systems, the comparison of the divergence and inability of various adaptive Kalman filtering methods are shown in Table II.

THE DIVERGENCE AND INABILITY RATES (%) OF NOISE COVARIANCE MATRIX ESTIMATION METHODS IN DIFFERENT SYSTEMS.

| Method | OB ($n_x=n_z$) | | | | OB ($n_x>n_z$) | | | | UO | | | |
|---|---|---|---|---|---|---|---|---|---|---|---|---|
| | $P_{d,Q}$ | $P_{d,R}$ | $P_{I,Q}$ | $P_{I,R}$ | $P_{d,Q}$ | $P_{d,R}$ | $P_{I,Q}$ | $P_{I,R}$ | $P_{d,Q}$ | $P_{d,R}$ | $P_{I,Q}$ | $P_{I,R}$ |
| Uncorrect | 83.1 | 79.6 | 0 | 0 | 85.7 | 51.3 | 0 | 0 | 83.2 | 85.9 | 0 | 0 |
| Rahul | 11.8 | 2.8 | 0 | 0 | 15.6 | 0.10 | 0.2 | 0.2 | 0 | 0 | 100 | 100 |
| Sage | 14.1 | 0 | 0 | 0 | 14.3 | 0.10 | 0 | 0 | 24.0 | 22.6 | 0 | 0 |
| ROSE | 1.80 | 0 | 0 | 0 | 1.50 | 0 | 0 | 0 | 2.0 | 0 | 0 | 0.1 |
| Maybeck | 32.8 | 30.0 | 0 | 0 | 89.2 | 54.5 | 0 | 0 | 0 | 2.10 | 13.6 | 13.6 |
| MDCM | 39.9 | 38.1 | 0 | 0 | 90.0 | 56.5 | 0 | 0 | 7.2 | 4.00 | 0 | 0 |
| Our | 0 | 0 | 0 | 0 | 0 | 0 | 0 | 0 | 0 | 0 | 0 | 0 |

Table II shows that the divergence rate and inability rate of the process and measurement noise covariance estimated by our method are the smallest in all systems. In the system where the measurement vector dimension is smaller than the state vector dimension, the divergence rates of the estimation results of Maybeck[14] and MDCM[19] are significantly greater than those of other methods, which confirms our summary of the correlation methods in the introduction. The inability rate of Rahul [20] in the UO system is 100%, which shows that the method cannot be applied to the UO system. This result proves that the closed-loop estimation scheme of estimation-calibration-correction proposed by us is effective.

Table I analyses the average estimation accuracy and standard deviation of adaptive Kalman filtering methods in L experiments. The results in Table II show that these L experiments include the divergence of the estimation results. However, under the premise of having a choice, researchers usually only choose the noise covariance estimation algorithm that converges to a specific system. Therefore, it is necessary to analyse the estimation performance of the estimation algorithm under the condition of convergence.

THE AVERAGE ESTIMATION ERROR AND STANDARD DEVIATION OF THE NOISE COVARIANCE MATRIX ESTIMATION METHOD IN DIFFERENT SYSTEMS.

| Method | OB ($n_x=n_z$) | | | | OB ($n_x>n_z$) | | | | UO | | | |
|---|---|---|---|---|---|---|---|---|---|---|---|---|
| | $\overline{\Delta Q}$ | $\overline{\Delta R}$ | $\sigma_{\Delta Q}$ | $\sigma_{\Delta R}$ | $\overline{\Delta Q}$ | $\overline{\Delta R}$ | $\sigma_{\Delta Q}$ | $\sigma_{\Delta R}$ | $\overline{\Delta Q}$ | $\overline{\Delta R}$ | $\sigma_{\Delta Q}$ | $\sigma_{\Delta R}$ |
| Uncorrect | 2.49 | 2.47 | 1.60 | 1.60 | 2.49 | 2.10 | 1.56 | 1.68 | 2.50 | 2.50 | 1.53 | 1.61 |
| Rahul | 1.49 | 0.54 | 1.70 | 0.66 | 2.49 | 0.25 | 1.57 | 0.00 | nan | nan | nan | nan |
| Sage | 2.00 | 0.25 | 1.63 | 0.11 | 1.96 | 0.19 | 1.60 | 0.28 | 2.20 | 0.41 | 1.79 | 1.02 |
| ROSE | **0.53** | 0.56 | 0.49 | 0.29 | 0.56 | 0.45 | 0.56 | 0.21 | 1.12 | 0.30 | 0.90 | **-0.01** |
| Maybeck | 2.43 | 2.28 | 1.69 | 1.63 | 2.62 | 2.19 | 1.56 | 1.65 | 0.48 | -0.08 | 0.21 | 0.58 |
| MDCM | 2.95 | 2.12 | 1.56 | 1.63 | 3.71 | 2.18 | 2.27 | 1.62 | 0.69 | **-0.40** | 0.69 | 0.30 |
| Our | 0.57 | **0.01** | **0.39** | **-0.17** | **0.55** | **-0.28** | **0.41** | **-0.51** | **0.27** | 0.31 | **0.16** | 0.07 |

We delete those divergent results in L experiments and then calculate the average estimation error and standard deviation and display the results in Table III. Table III shows that when the estimation results converge, the covariance matching methods such as Sage [16] and ROSE [31] have smaller estimation errors and standard deviations than other methods.

Our method can achieve similar performance to Sage and ROSE.

*3). The system state estimation error is defined as:*

$$\begin{cases} \Delta y_{j,k} = | y_{true,j,k} - H_{j,k} x_{j,k/k} | \\ error_j = (\sum_{k=1}^{k=T_m} \Delta y_{j,k}^2)^{0.5} \quad (j=1,2,...,L) \end{cases} \quad (45)$$

$\Delta y_{j,k}$ represents the system state estimation error at time instant $k$ of the $j$th experiment, and $error_j$ represents the average system state estimation error of the $j$th experiment. $y_{true,j,k}$ is the true measurement vector generated by Algorithm 2, and $H_{j,k}$ is the observation matrix generated by Algorithm 1.

In the UO system and the OB system, the system state estimation error $error_j$ of the NC2 adaptive method in L experiments is shown in Figure 4.

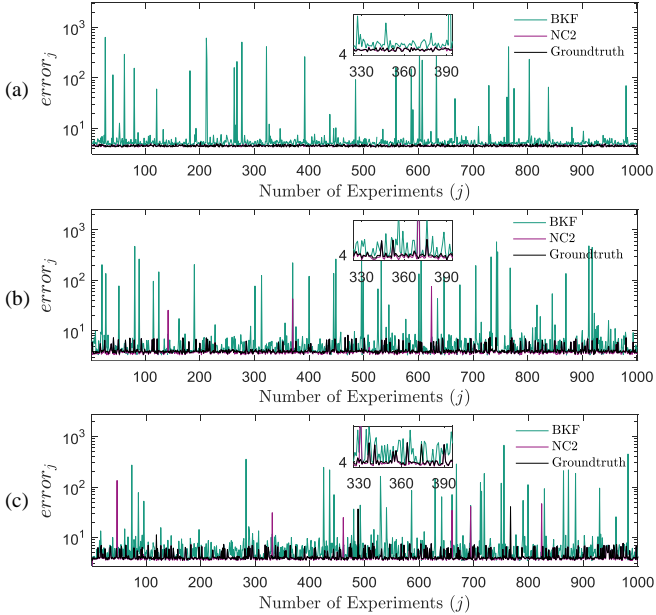

Fig. 4. State estimation error of the Kalman filter in the noise time-varying system (a), OB system and $n_{z=} n_x$ (b), OB system and $n_{z<} n_x$ (c), UO system.

In Figure 4, "BKF" and "NC2" represent the average system state estimation error of the benchmark Kalman filter method and our method when the initial process and measurement noise covariance matrices deviate from the true value. "Groundtruth" represents the average system state estimation error of the Kalman filter method when the process and measurement noise covariance matrices are the same as the true value. We regard "Groundtruth" as the best state estimation accuracy that the Kalman filter can achieve in a randomly generated system. Figure 4 shows that in 3000 random experiments of the three types of systems mentioned above, the system state estimation error of our method is significantly lower than that of the benchmark Kalman filter (BKF). This result proves that our method effectively improves the system state estimation accuracy of the Kalman filter.

### D. Application of NC2 in MOT

We evaluate the effectiveness of the NC2 method in KITTI MOT datasets [29], which provide LiDAR point clouds and 3D bounding box trajectories. On the KITTI validation set, we use the detection of the 3D clustering method [28] as the input of the tracker. On the KITTI test set, we use the detection of Voxel R-CNN [30] as the input of the tracker.

*a). KITTI MOT val dataset*

On the KITTI validation dataset, we analyse the estimation performance of adaptive Kalman filtering methods. Figure 5 shows the estimated noise covariance matrix of a series of tracking sequences. Each point in the figure represents the F-norm average of a tracking sequence's estimated process and measurement noise covariance matrices. The initial noise covariance matrix is the random element matrix generated by (39).

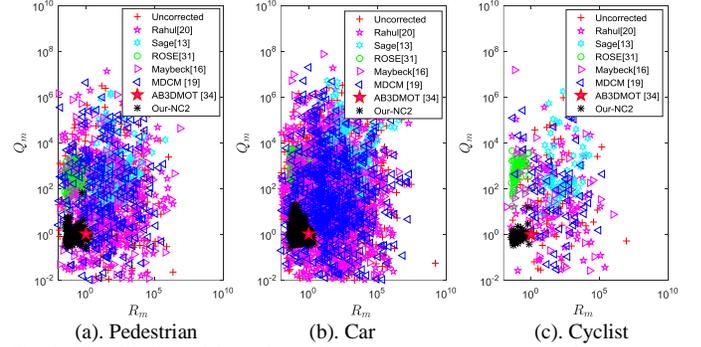

(a). Pedestrian (b). Car (c). Cyclist
Fig. 5. Visualization of the estimated process and measurement noise covariance matrix on the KITTI validation dataset.

In Fig. 5, the symbol "★" represents the process and measurement noise covariance matrix given in AB3DMOT [21]. Compared with other adaptive Kalman filtering methods, the divergence and variance of our method are the smallest.

To evaluate the effectiveness of the NC2 method in improving target tracking performance, we use the AB3DMOT method as a benchmark for comparison. For a fair comparison, we use the same point cloud clustering results, object matching method, and life cycle management method as AB3DMOT. In the tracking process, AB3DMOT uses the Kalman filter and a fixed noise covariance matrix. We use the noise covariance estimated online by the NC2 filter. The common metrics (MOTP, MOTA, AMOTP, AMOTA, sAMOTA, ML (mostly lost), IDS, and Flag) of multi-object tracking [21] are used for evaluation. Table IV shows the comparison between AB3DMOT and our method in the KITTI MOT validation dataset.

3D MULTIOBJECT TRACKING PERFORMANCE ON THE KITTI-MOTS [29] VALIDATION SET

| Method | Car | | Pedestrian | | Cyclist | |
|--------|---------|-------|----------|-------|---------|-------|
|        | AB3D[21] | Our  | AB3D[21] | Our   | AB3D[21] | Our  |
| MOTP   | 61.38   | **64.11** | 46.27 | **50.34** | 50.75 | **56.64** |
| MOTA   | 85.99   | **86.18** | 53.62 | **56.45** | 77.23 | **77.30** |
| AMOTP  | 59.32   | **61.79** | 31.75 | **35.64** | 47.31 | **57.32** |
| AMOTA  | **45.36** | 45.22 | 20.57 | **21.23** | 38.83 | **39.38** |
| sAMOTA | **93.23** | 93.03 | 59.58 | **61.60** | 84.52 | **85.37** |
| ML     | 3.78    | **3.24** | 23.94 | **20.42** | 7.14 | **3.57** |
| IDS    | 0       | 0     | 35    | **28**  | 2     | 2     |
| Flag   | **26**  | 30    | 380   | **298** | 34    | **27** |

We use bold red to emphasize the best performance value. The results in Table IV show that our method has improved on most indicators compared to AB3DMOT. This result proves that the adaptive filter with noise covariance matrix estimation ability proposed in this paper performs better than the Kalman filter in actual tracking applications. Our method does not significantly improve vehicle tracking performance compared with pedestrians and bicycles. After analysis, we found that in the point cloud clustering results, the detection accuracy of vehicles is much higher than that of pedestrians and bicycles. Therefore, the measurement noise of the vehicle is slight, and the amplitude of the change over time is small. In other words, most of the time, the measurement noise covariance of the vehicle in the filter can be regarded as a fixed value. Therefore, adaptive noise covariance estimation does not improve the tracking performance of vehicles as much as pedestrians and bicycles.

*b). KITTI MOT test dataset*

Pedestrian tracking has attracted much attention due to its vast potential in commercial applications. Tracking pedestrians is usually more difficult than tracking vehicles, as confirmed in Table IV. The main reason is that pedestrians are small in size and changeable in exercise patterns. A smaller body size makes the measurement noise of the laser on pedestrians increase, and the changeable motion mode makes the process noise of the filter change frequently. Table 4 shows that the NC2 method can improve the tracking performance of pedestrians when using 3D clustering results as the tracking input. In the KITTI MOT test dataset, we use the deep point cloud detection method (Voxel R-CNN [30]) to detect pedestrians and use the NC2 method for tracking. To increase credibility, we upload the tracking result of the NC2 method to the official KITTI server for performance evaluation and compared with the current state-of-the-art pedestrian tracking methods (QuasiDense [32], CenterTrack [33], EagerMOT [34], and AB3DMOT [21]). Table 5 shows the quantification of multi-pedestrian tracking results on the KITTI MOT test dataset. KITTI's official performance evaluation metric (HOTA) is used for quantitative evaluation.

PERFORMANCE COMPARISON OF THE STATE-OF-THE-ART LIDAR POINT CLOUD MULTIOBJECT TRACKING METHODS ON THE KITTI 2D MOT LEADERBOARD.

| Method | | Input data | HOTA | DetPr | LocA | MOTA | Time |
|---|---|---|---|---|---|---|---|
| QuasiDense | | 2D | 41.12 | **70.39** | **77.87** | 55.55 | 14.3 |
| CenterTrack | | 2D | 40.35 | 66.83 | 77.81 | 53.84 | **222** |
| EagerMOT | | 2D+3D | 39.38 | 61.49 | 71.25 | 49.82 | 90.9 |
| AB3D | | 3D | 35.57 | 60.31 | 71.26 | 38.93 | 212 |
| Ours | NC2 | 3D | 44.30 | 52.43 | 72.08 | 44.18 | 91.6 |
| | FNC2 | 2D+3D | **46.55** | 59.38 | 72.07 | **56.05** | 91.6 |

The data in Table 5 come from the pedestrian multi-object tracking leaderboard on KITTI's official website (http://www.cvlibs.net/datasets/kitti/eval_tracking.php). Be-Track and EagerMOT are the top 3 LIDAR point cloud multiobject tracking methods in the leaderboard. Table 5 shows that, compared with the AB3DMOT method, the tracking performance of our method is significantly improved in most indicators except detection accuracy and computational efficiency (excluding the 3D detector part). The detection accuracy depends on the performance of the point cloud detection algorithm. The tracking results of the NC2 group use Voxel R-CNN [30] open-source code for point cloud detection, and its detection accuracy is lower than other methods. To compare the tracking performance fairly, we use LIDAR and visual fusion technology to optimize the detection results of Voxel R-CNN. We named the method that takes the detection result optimized by multisensor fusion technology as the input as FNC2. Table 5 shows that the detection accuracy of FNC2 is similar to that of AB3DMOT, but its tracking performance is significantly better than that of the other methods. When the NC2 method was first submitted to the KITTI server for evaluation, its tracking performance surpassed all methods that use lidar for pedestrian tracking. These results once again prove the feasibility of the filter with adaptive noise covariance estimation capability proposed in this paper in practical applications.

## VII. CONCLUSION

This paper proposes an estimation-calibration-correction closed-loop estimation method (NC2) to adaptively estimate the process and measurement noise covariance matrices and system state of linear discrete noise time-varying systems. The estimation method's unbiased and probabilistic convergence is proven under the additional assumptions of detectable and Gaussian noise. This work differs from the existing adaptive covariance matrix estimation algorithms. First, there are fewer restrictions on technical assumptions. In addition to observable systems, this method can also be applied to unobservable systems. Second, there is no strict restriction on the dimensions of the state transition matrix and the observation matrix of the filter. Finally, the process and measurement noise covariance matrices can be estimated simultaneously. The simulation results prove the applicability and convergence of the method in multiple systems. The experimental results on the KITTI MOT dataset prove the feasibility of the method in practical applications. In the future, we plan to further improve this method's computational efficiency and accuracy under the premise of ensuring stability.

## APPENDIX A

We carried out multiobject tracking experiments on both the KITTI tracking set and our intelligent vehicle platform, and the experimental videos were released on GitHub (https://github.com/JiangChao2009/NC2-MOT).

## APPENDIX B. PROOF OF THEOREM 1

### A. Proof of Property 1 in Theorem 1

*Proof:* According to the definitions of equations (18) and (26), $E_{A,max,k}$ is the maximum or minimum value of the $\tau_{A,k}(\eta)$ sequence. Therefore, as long as it is proven that when $\delta_\varpi = \delta_v$, the expectation of $\tau_{A,k}(\eta)$ is 0, and $\tau_{A,k}(\eta)$ converges to its own expectation, then it can be proved that $E_{A,max,k}$ converges to 0 when $\delta_\varpi = \delta_v$.

The expectation of $\tau_{A,k}(\eta)$ is:

$$E(\tau_{A,k}(\eta)) = E(\frac{\sum_{t=k-\eta+i}^{k}\tau_{I,t}\tau_{I,t-i}^T}{\eta-i} - \frac{\sum_{t=k-\eta+i}^{k}\tau_{I,t}\sum_{t=k-\eta+i}^{k}\tau_{I,t-i}}{(\eta-i+1)(\eta-i)})$$
$$= E(\tau_{I,k}\tau_{I,k-i}^T) - E(\tau_{I,k})E(\tau_{I,k-i}) \qquad (46)$$
$$= E(\tau_{I,k}\tau_{I,k-i}^T) - E(H_k(x_k - x_{k|k-1}) + v_k)E(\tau_{I,k-i})$$

In Kalman filtering that meets the basic assumptions, $x_{k|k-1}$ is an unbiased estimate of $x_k$, that is, $E(\tau_{I,k})=0$. Therefore:

$$E(\tau_{A,k,i}(\eta)) = E(\tau_{I,k}\tau_{I,k-i}^T) \qquad (47)$$

That is, $\tau_{A,k,i}(\eta)$ is an unbiased estimate of $E(\tau_{I,k}\tau_{I,k-i}^T)$.

Let $n_{k,k-i} = E(\tau_{I,k}\tau_{I,k-i}^T)$, $\hat{x}_{k|k-1} = x_k - x_{k|k-1}$, then:

$$\begin{aligned} n_{k,k-i} &= E((H_k\hat{x}_{k|k-1} + v_k)(H_{k-i}\hat{x}_{k-i|k-i-1} + v_{k-i})^T) \\ &= H_k E(\hat{x}_{k|k-1}\hat{x}_{k-i|k-i-1}^T)H_{k-i}^T + H_k E(\hat{x}_{k|k-1}v_{k-i}^T) \\ &\quad + E(v_k\hat{x}_{k-i|k-i-1}^T)H_{k-i}^T + E(v_kv_{k-i}^T) \end{aligned} \qquad (48)$$

Under *Assumption 2.4*, the measurement noise sequence $\{v_k\}$ is Gaussian white noise with zero mean, and $v_k$ and $\hat{x}_{k-i|k-i-1}$ at $k-i$ ($i>0$) are also uncorrelated. therefore:

$$n_{k,k-i} = H_k E(\hat{x}_{k|k-1}\hat{x}_{k-i|k-i-1}^T)H_{k-i}^T + H_k E(\hat{x}_{k|k-1}v_{k-i}^T) \qquad (49)$$

$$\begin{aligned} \hat{x}_{k|k-1} &= x_k - x_{k|k-1} = \Phi_{k-1}x_{k-1} + \varpi_{k-1} - \Phi_{k-1}x_{k-1|k-1} \\ &= \Phi_{k-1}(I - \tilde{K}_{k-1}H_{k-1})x_{k-1|k-2} - \Phi_{k-1}\tilde{K}_{k-1}v_{k-1} + \varpi_{k-1} \end{aligned} \qquad (50)$$

where $K_{k-1}$ is the Kalman gain, and iteratively solving (50) can be obtained:

$$\hat{x}_{k-i+1|k-i} = \Phi_{k-i}(I - K_{k-i}H_{k-i})\hat{x}_{k-i|k-i-1} - \Phi_{k-i}K_{k-i}v_{k-i} + \varpi_{k-i} \qquad (51)$$

Let $\tilde{B}_j = I - K_jH_j$, $\hat{x}_{k|k-1}$ can be expressed as:

$$\begin{aligned} \hat{x}_{k|k-1} = \prod_{j=k-1}^{k-i}\Phi_j\tilde{B}_j\hat{x}_{k-i|k-i-1} - \sum_{m=2}^{i}\prod_{j=k-1}^{k-m+1}\Phi_j\tilde{B}_j\Phi_{k-m}K_{k-m}W_{k-m} \\ -\Phi_{k-1}K_{k-1}W_{k-1} + \sum_{m=2}^{i}\prod_{j=k-1}^{k-m+1}\Phi_j\tilde{B}_j\varpi_{k-m} + \varpi_{k-1} \end{aligned} \qquad (52)$$

Since $v_k$, $\varpi_{k-m}$ ($m \in [1,2,\ldots i]$) and $\hat{x}_{k-i+1|k-i}$ are uncorrelated to each other, so:

$$E(\hat{x}_{k|k-1}\hat{x}_{k-i|k-i-1}^T) = \prod_{j=k-1}^{k-i}(\Phi_j\tilde{B}_j)E(\hat{x}_{k-i|k-i-1}\hat{x}_{k-i|k-i-1}^T) = \prod_{j=k-1}^{k-i}(\Phi_j\tilde{B}_j)P_{k-i|k-i-1} \quad (53)$$

$$\begin{aligned} E(\hat{x}_{k|k-1}v_{k-i}^T) &= -\prod_{j=k-1}^{k-i+1}(\Phi_j\tilde{B}_j)\Phi_{k-i}\tilde{K}_{k-i}E(v_{k-i}v_{k-i}^T) \\ &= -\prod_{j=k-1}^{k-i+1}(\Phi_j\tilde{B}_j)\Phi_{k-i}\tilde{K}_{k-i}R_{k-i} \end{aligned} \qquad (54)$$

Substituting (53) and (54) into (49), we can get:

$$\begin{aligned} n_{k,k-i} &= H_k\prod_{j=k-1}^{k-i}\Phi_j\tilde{B}_jP_{k-i|k-i-1}H_{k-i}^T - H_k\prod_{j=k-1}^{k-i+1}\Phi_j\tilde{B}_j\Phi_{k-i}\tilde{K}_{k-i}R_{k-i} \\ &= H_k\prod_{j=k-1}^{k-i+1}\Phi_j\tilde{B}_j\Phi_{k-i}(\tilde{B}_{k-i}P_{k-i|k-i-1}H_{k-i}^T - \tilde{K}_{k-i}R_{k-i}) \\ &= H_k\prod_{j=k-1}^{k-i+1}\Phi_j\tilde{B}_j\Phi_{k-i}((I-\tilde{K}_{k-i}H_{k-i})P_{k-i|k-i-1}H_{k-i}^T - \tilde{K}_{k-i}R_{k-i}) \\ &= H_k\prod_{j=k-1}^{k-i+1}\Phi_j\tilde{B}_j\Phi_{k-i}(P_{k-i|k-i-1}H_{k-i}^T - \tilde{K}_{k-i}(H_{k-i}P_{k-i|k-i-1}H_{k-i}^T + R_{k-i})) \\ &= H_k\prod_{j=k-1}^{k-i+1}\Phi_j\delta_\varpi R_j(M_{j-1} + H_j\delta_vQ_{j-1}H_j^T + \delta_\varpi R_j)^{-1}\Phi_{k-i}H_{k-i}^{-1} \\ &\quad \times((M_{k-i+1} + H_{k-i}Q_{k-i-1}H_{k-i}^T) - (M_{k-i+1} + \delta_vH_{k-i}Q_{k-i-1}H_{k-i}^T)(M_{k-i+1} \\ &\quad + \delta_vH_{k-i}Q_{k-i-1}H_{k-i}^T + \delta_\varpi R_{k-i})^{-1}(M_{k-i+1} + H_{k-i}Q_{k-i-1}H_{k-i}^T + R_{k-i})) \end{aligned} \qquad (55)$$

where, $M_{j-1} = H_j\Phi_{j-1}P_{j-1|j-1}\Phi_{j-1}^TH_j^T$.

When $i>0$ and $\delta_\varpi = \delta_v = \delta$, especially when $\delta = 1$, according to (55), we can obtain:

$$E(\tau_{A,k,i}(\eta)) = E(\tau_{I,k}\tau_{I,k-i}^T) = n_{k,k-i} = 0 \qquad (56)$$

That is, the innovation sequence $\{\tau_{I,k}\}$ is a sequence of independent and identically distributed random variables. According to the law of large numbers [35], we can obtain that the sample mean $\bar{\tau}_{I,k}$ converges to expectation $\mu_\tau = E[\tau_{I,k}] = 0$ in probability.

Based on the above conclusions, we can prove that $\tau_{A,k}(\eta)$ converges to its own expectation.

$$\begin{aligned} \tau_{A,k}(\eta) &= \frac{1}{\eta-i}\sum_{t=k-\eta+i}^{k}(\tau_{I,k,t} - \bar{\tau}_{I,k})(\tau_{I,k,t-i} - \bar{\tau}_{I,k})^T \\ &= \frac{1}{\eta-i}\sum_{t=k-\eta+i}^{k}(\tau_{I,k,t}\tau_{I,k,t-i} - \bar{\tau}_{I,k}(\tau_{I,k,t} + \tau_{I,k,t-i}) + \bar{\tau}_{I,k}^2) \\ &= \frac{1}{\eta-i}\sum_{t=k-\eta+i}^{k}(\tau_{I,k,t}\tau_{I,k,t-i}) + o(1) \xrightarrow{\eta\to\infty} E(\tau_{I,k}\tau_{I,k-i}^T) = 0 \end{aligned} \qquad (57)$$

That is, when $\delta_\varpi = \delta_v = \delta$, including $\delta = 1$, according to the law of large numbers [35], $\tau_{A,k}(\eta)$ converges to 0 in probability. Therefore, according to the definitions of (18) and (26), we can prove that $E_{A,max,k}$ converges to 0 when $\delta_\varpi = \delta_v$.

B. *Proof of Property 2 in Theorem 1.*

*proof:* In (55), let:

$$\begin{aligned} f_1(\delta_\varpi, \delta_v) &= \prod_{j=k-1}^{k-i+1}\Phi_j\delta_\varpi R_j(M_{j-1} + H_j\delta_vQ_{j-1}H_j^T + \delta_\varpi R_j)^{-1} \\ f_2(\delta_\varpi, \delta_v) &= \frac{-(M_{k-i+1} + \delta_vH_{k-i}Q_{k-i-1}H_{k-i}^T)}{M_{k-i+1} + \delta_vH_{k-i}Q_{k-i-1}H_{k-i}^T + \delta_\varpi R_{k-i}} \end{aligned} \qquad (58)$$

Then:

$$\begin{aligned} n_{k,k-i} &= H_kf_1(\delta_\varpi, \delta_v)\Phi_{k-i}H_{k-i}^{-1}((M_{k-i+1} + H_{k-i}Q_{k-i-1}H_{k-i}^T) \\ &\quad + f_2(\delta_\varpi, \delta_v)(M_{k-i+1} + H_{k-i}Q_{k-i-1}H_{k-i}^T + R_{k-i})) \end{aligned} \qquad (59)$$

When $Q_{k-i-1}$ is a positive semidefinite matrix, for any $\delta_{\varpi 0}$ and $\delta_{\varpi 1}$ satisfying the condition $0 < \delta_{\varpi 0} < \delta_{\varpi 1}$, there are:

$$\frac{f_1(\delta_{\varpi 0}, \delta_v)}{f_1(\delta_{\varpi 1}, \delta_v)} = \prod_{j=k-1}^{k-i+1}(\frac{\delta_{\varpi 0}(M_{j-1} + H_j\delta_vQ_{j-1}H_j^T) + \delta_{\varpi 0}\delta_{\varpi 1}R_j}{\delta_{\varpi 1}(M_{j-1} + H_j\delta_vQ_{j-1}H_j^T) + \delta_{\varpi 1}\delta_{\varpi 0}R_j}) < 1$$

$$\begin{aligned} f_2(\delta_{\varpi 0}, \delta_v)^{-1} - f_2(\delta_{\varpi 1}, \delta_v)^{-1} \\ = (M_{k-i+1} + \delta_vH_{k-i}Q_{k-i-1}H_{k-i}^T)^{-1}((\delta_{\varpi 1} - \delta_{\varpi 0})R_{k-i}) > 0 \end{aligned}$$

That is, both $f_1$ and $f_2$ are monotonically increasing functions of $\delta_\varpi$.

When $Q_{k-i-1}$ and $R_{k-i}$ are positive semidefinite, for any $0 < \delta_{v0} < \delta_{v1}$, there are:

$$\frac{f_1(\delta_\varpi, \delta_{v0})}{f_1(\delta_\varpi, \delta_{v1})} = \prod_{j=k-1}^{k-i+1}(\frac{M_{j-1} + H_j\delta_{v1}Q_{j-1}H_j^T + \delta_\varpi R_j}{M_{j-1} + H_j\delta_{v0}Q_{j-1}H_j^T + \delta_\varpi R_j}) > 1$$

$$\begin{aligned} &f_2(\delta_\varpi, \delta_{v0})^{-1} - f_2(\delta_\varpi, \delta_{v1})^{-1} \\ &= \frac{(\delta_{v0} - \delta_{v1})\delta_\varpi R_{k-i}H_{k-i}Q_{k-i-1}H_{k-i}^T}{(M_{k-i+1} + \delta_{v1}H_{k-i}Q_{k-i-1}H_{k-i}^T)(M_{k-i+1} + \delta_{v0}H_{k-i}Q_{k-i-1}H_{k-i}^T)} < 0 \end{aligned}$$

That is, both $f_1$ and $f_2$ are monotonically decreasing functions of $\delta_v$.

In (58), $r_{k,k-i}$ is an increasing function of $f_1$ and $f_2$, so $r_{k,k-i}$ is a monotonically increasing function of $\delta_\varpi$ and a monotonically decreasing function of $\delta_v$. Combining the conclusion that $r_{k,k-i}=0$ when $\delta_\varpi=\delta_v=1$, we can obtain $r_{k,k-i}>0$ when $\delta_\varpi>1 \& \delta_v\leq1$ or $\delta_\varpi\geq1 \& \delta_v<1$. In (57), $\tau_{A,k}(\eta)$ converges to $r_{k,k-i}$ in probability, so when $r_{k,k-i}>0$, $\lim_{\eta\to\infty}\tau_{A,k}(\eta)>0$. Similarly, when $\delta_\varpi<1 \& \delta_v\geq1$ or $\delta_\varpi\leq1 \& \delta_v>1$, $r_{k,k-i}<0$, $\lim_{\eta\to\infty}\tau_{A,k}(\eta)<0$. Obviously the following conclusions are correct:

$$\max(\lim_{\eta\to\infty}\tau_{A,k}(\eta))>0, \quad \min(\lim_{\eta\to\infty}\tau_{A,k}(\eta))<0.$$

That is, on the premise that $\eta$ is large enough, $E_{A,max,k}>0$ when $\delta_\varpi>1 \& \delta_v\leq1$ or $\delta_\varpi\geq1 \& \delta_v<1$; $E_{A,max,k}<0$ when $\delta_\varpi<1 \& \delta_v\geq1$ or $\delta_\varpi\leq1 \& \delta_v>1$.

### C. Proof of Property 3 in Theorem 1.

*proof:* According to the definitions of equations (19) and (26), $E_{G,max,k}$ is the maximum or minimum element value of the $E_{G,k}$ vector. Therefore, as long as it is proven that $E_{G,k}$ converges to 0 when $\delta_\varpi=\delta_v=1$, it can be proven that $E_{G,max,k}$ also converges to 0 when $\delta_\varpi=\delta_v=1$.

In (23), $\mathrm{E}(\hat{\mu}_{|\tau|}) = \sum_{t_k=k-N+1}^{k}\mathrm{E}(|\tau_{I,k}|)/N = \mu_{|\tau|}$, so the sample mean $\hat{\mu}_{|\tau|}$ is an unbiased estimate of expected $\mu_{|\tau|}$. In addition, the innovation sequence $\{\tau_{I,k}\}$ is independent and identically distributed when $\delta_\varpi=\delta_v=1$. According to the definition of sequence correlation [36], $\{\hat{\mu}_{|\tau|}\}$ is also an independent and identically distributed random variable sequence when $\delta_\varpi=\delta_v=1$. According to the law of large numbers [35], when $\delta_\varpi=\delta_v=1$, $\hat{\mu}_{|\tau|}$ converges to the expectation $\mu_{|\tau|}$ in probability. Moreover, when $\delta_\varpi=\delta_v=1$, according to (24), $\hat{T}_{g,k}=T_{g,k}$. So when $\delta_\varpi=\delta_v=1$, we can get:

$$\begin{cases} \forall \varepsilon>0, \lim_{k\to\infty}\mathrm{P}\{|P_{e,k}-P_{g,k}|>\varepsilon\}=0 \\ \lim_{k\to\infty}\sqrt{P_{e,k}^2+P_{g,k}^2}=1/\sqrt{\pi} \end{cases} \quad (60)$$

Substituting (61) into (19), we can get:

$$\forall \varepsilon>0, \lim_{k\to\infty}\mathrm{P}\{|E_{G,k}|>\varepsilon\}=0 \quad (61)$$

That is, $E_{G,k}$ converges to 0 in probability, so $E_{G,max,k}$ converges to 0 in probability when $\delta_\varpi=\delta_v=1$.

### D. Proof of Property 4 in Theorem 1.

*proof:* The innovation covariance can be expressed as:

$$S_k = H_k(\Phi_{k-1}P_{k-1|k-1}\Phi_{k-1}^T + \delta_{\upsilon,k}\sum_{i=1}^{n_x}\sum_{j=1}^{n_x}Q_{k-1,i,j}D_{e,\hat{Q}_{k-1}})H_k^T + \delta_{\varpi,k}\sum_{i=1}^{n_z}\sum_{j=1}^{n_z}R_{k,i,j}D_{e,\hat{R}_k}$$

Therefore, in Equation (25), $P_{e,k}(\delta_\varpi, \delta_v)$ is the monotonic decreasing function of $\delta_\varpi$ and $\delta_v$.

Substituting (25) into (19), we can obtain that $E_{G,k}$ is a monotonically decreasing function of $\delta_\varpi$ and $\delta_v$. Combined with the conclusion that $E_{G,k}$ converges to 0 when $\delta_\varpi=\delta_v=1$, we can obtain that when $\delta_\varpi<1 \& \delta_v\leq1$ or $\delta_\varpi\leq1 \& \delta_v<1$:

$$E_{G,k} = (P_{e,k}-P_{g,k})/\sqrt{(P_{e,k}^2+P_{g,k}^2)} > 0$$

When $\delta_\varpi>1 \& \delta_v\geq1$ or $\delta_\varpi\geq1 \& \delta_v>1$:

$$E_{G,k} = (P_{e,k}-P_{g,k})/\sqrt{(P_{e,k}^2+P_{g,k}^2)} < 0$$

Moreover:

$$\lim_{\substack{\delta_\varpi\to\infty \\ \delta_v\to\infty}} E_{G,k} = -1, \quad \lim_{\substack{\delta_\varpi\to 0 \\ \delta_v\to 0}} E_{G,k} \approx 1$$

From the above results, it can be concluded that the larger the deviation between the true noise covariance matrix and the estimated noise covariance matrix, the larger the modulus of $E_{G,k}$, and the maximum value is 1. $E_{G,max,k}$ is the element with the largest module in the $E_{G,k}$ vector, so it also has the above properties.

Prove up.

## APPENDIX C. GENERATION METHOD OF SYSTEM PARAMETERS AND MEASUREMENT SEQUENCE

### A. System parameter generation method

Algorithm 1 is used to generate random system parameters, including the state transition matrix $\Phi$, the observation matrix $H$, the process noise covariance matrix $Q$, and the measurement noise covariance matrix $R$.

---
**Algorithm 1: Generate system parameters**
---
**Input :** $n_x, n_z, con$
**Output :** $\Phi, H, Q, R$
**for** $j=1$ **to** L, **do**
    **while** $con$, **do**    #Condition
        #Generate random parameters
        $\Phi_c = \mathbf{round}(c_1 * c_2^{(\mathbf{randn}(n_x,n_x)+c_3)})/c_1$
        $H_c = \mathbf{round}(c_1 * c_2^{(\mathbf{randn}(n_z,n_x)+c_3)})/c_1$
        $Q_c = c_4 * \mathbf{abs}(\mathbf{randn}(n_x,n_x))$
        $R_c = c_4 * \mathbf{abs}(\mathbf{randn}(n_z,n_z))$
        #Calculate positive semidefiniteness
        $pd = \mathbf{ED}(Q_c) \& \mathbf{ED}(R_c)$
        #Calculate observability
        $ob = \mathbf{OBC}(n_x, n_z, \Phi_c, H_c)$
        #Calculate detectability
        $de = \mathbf{PBH}(n_x, \Phi_c, H_c)$
    **end**
    $\Phi(j)=\Phi_c, H(j)=H_c, Q(j)=Q_c, R(j)=R_c$
**end for**

---

where $c_1=100$, $c_2=2$, $c_4=5$, and L is the number of experiments for each class of systems (L = 1000). If $c_3=-10$ and $con=(pd\neq1 \mid de\neq1 \mid ob\neq1)$, an observable system can be generated. If $c_3=10$, $con=(pd\neq1 \mid de\neq1 \mid ob\neq0)$ can generate a detectable but unobservable system. In addition, the dimensions of the generated measurement vector and state vector can be controlled by $n_x$ and $n_z$. For example, if $n_x=n_z=3$, the same dimension's measurement vector and state vector will be generated; if $n_x=3$, $n_z=2$, the dimension of the generated measurement direction will be smaller than the dimension of the state vector.

In the **while** loop of generating random parameters, only if $con$ is not 0 will it jump out of the loop. **round** is the integer function, and **randn** is the random number generating function. **ED** is the eigenvalue discriminant function [37] used to judge whether the matrix is semidefinite. OBC is an observability discrimination function [38] used to judge whether the generated system is observable. **PBH** is a rank discriminant function [39] used to judge whether the generated system is detectable. *pd*, *ob* and *de* are all bool variables. If *pd* is 1, then *Q* and *R* are positive semidefinite; otherwise, *Q* or *R* are not positive semidefinite. If *ob* is 1, the system is observable; otherwise, the system is not observable. If *de* is 1, the system can be detected; otherwise, the system cannot be detected.

*B. Measurement sequence generation method*

The measurement sequence with time-varying noise covariance is generated using Algorithm 2.

---
Algorithm 2: Generate a series of measurements with time-varying noise
---
**Input** : $\Phi, H, Q, R, L, T_m, n_x, n_z$
**Output** : $y_m, y_{true}, Q_{true}, R_{true}$
**for** $j = 1$ **to** L, **do**
　　$c_r = \mathbf{abs}(2*\mathbf{randn}(3,2))$ , $x = \mathbf{zeros}(n_x)$
　　**for** $k = 1$ **to** $T_m$, **do**
　　　　# Generate time-varying noise covariance matrix
　　　　$Q_{true}(j,k) = c_r * (\mathbf{round}(3*k/T_m),1) * Q(j)$
　　　　$R_{true}(j,k) = c_r * (\mathbf{round}(3*k/T_m),2) * R(j)$
　　　　# Generate time-varying process noise
　　　　$v = \mathbf{randn}(1,n_x) * \mathbf{chol}(Q_{true}(j,k))$
　　　　# Generate system states with time-varying noise
　　　　$x = \Phi(j) * x + v'$
　　　　# Generate time-varying measurement noise
　　　　$w = \mathbf{randn}(1,n_z) * \mathbf{chol}(R_{true}(j,k))$
　　　　# Generate measurements with time-varying noise
　　　　$y = H(j) * x + w'$
　　　　$y_m(:,k,j) = y$ , $y_{true}(:,k,j) = H(j) * x$
　　**end for**
**end for**

In Algorithm 2, $T_m$ is the length of the measurement sequence ($T_m = 900$). $y_m$ and $y_{true}$ are $L \times T_m$ dimensional matrices. $y_m$ stores the measurement sequence with time-varying noise, and $y_{true}$ stores the measurement sequence without noise. $Q_{true}$ and $R_{true}$ store the true process and measurement noise covariance matrices, respectively. **abs** is a function of absolute value.